\documentclass{article}

\usepackage[final, nonatbib]{nips_2018}

\usepackage[utf8]{inputenc} 
\usepackage[T1]{fontenc}    
\usepackage{hyperref}       
\usepackage{url}            
\usepackage{booktabs}       
\usepackage{amsfonts}       
\usepackage{nicefrac}       
\usepackage{microtype}      

\usepackage[compact]{titlesec}
\titlespacing{\section}{0pt}{*0}{*0}
\titlespacing{\subsection}{0pt}{*0}{*0}
\titlespacing{\subsubsection}{0pt}{*0}{*0}

\usepackage{subcaption}
\usepackage{amsmath}
\usepackage{amssymb}
\usepackage[pdftex]{graphicx}
\usepackage{multirow}

\usepackage{textgreek}

\title{Bayesian Nonparametrics for Non-exhaustive Learning}

\author{
  Yicheng Cheng \\
  Computer \& Information \\
  Science \\
  IUPUI \\
  \texttt{chengyic@iu.edu} \\
  \And
  Bartek Rajwa \\
  Bindley Bioscience Center \\
  Purdue University \\
  \texttt{rajwa@cyto.purdue.edu} \\
  \And
  Murat Dundar \\
  Computer \& Information \\
  Science \\
  IUPUI \\
  \texttt{mdundar@iupui.edu} \\
}

\begin{document}

\maketitle

\begin{abstract}
Non-exhaustive learning (NEL) is an emerging machine-learning paradigm designed to confront the challenge of non-stationary environments characterized by a non-exhaustive training sets lacking full information about the available classes. Unlike traditional supervised learning that relies on fixed models,
NEL utilizes self-adjusting machine learning to better accommodate the non-stationary nature of the real-world problem, which is at the root of many recently discovered limitations of deep learning.
Some of these hurdles led to a surge of interest in several research areas relevant to NEL such as open set classification or zero-shot learning. The presented study which has been motivated by two important applications proposes a NEL algorithm built on a highly flexible, doubly non-parametric Bayesian Gaussian mixture model that can grow arbitrarily large in terms of the number of classes and their components. We report several experiments that demonstrate the promising performance of the introduced model for NEL.
\end{abstract}

\section{Introduction}
Many contemporary data science problems originate in non-stationary environments where new classes of patterns can emerge at any time. This difficulty creates an ill-defined setting for traditional classification algorithms that take stationarity for granted and employing fixed models trained with a presumably exhaustive set of classes. Under these circumstances, samples from classes not represented in the training set are typically misclassified into one of the observed categories. This outcome creates a two-sided problem. First, the unknown class, which could potentially represent a significant abnormality such as a residual population of cancer cells in bone marrow \cite{rajwa2017automated} or an emerging pathogenic bacteria strain present in food products \cite{akova2010machine}, cannot be appropriately detected leading to potentially catastrophic consequences. Second, even if the unknown classes do not have any significance, misclassifying irrelevant samples into classes of practical importance raises doubts about the overall stability of the machine-learning systems, as has recently been the case with some well-established deep learning models \cite{moosavi2016deepfool,nguyen2015deep}. 


NEL is closely related to zero-shot learning (ZSL) \cite{palatucci2009zero} and open-set classification (OSC) \cite{openset}. The differences among these three approaches can be clarified by describing the treatment and vocabulary of the problematic classes. In the ZSL literature classes that are known and represented in the training dataset are termed \textit{seen} classes. The OSC literature uses the term \textit{known knowns}. Further, some classes are known, but they are not represented during training. In the ZSL literature these are called \textit{unseen} classes, and \textit{known unknowns} in the OSC. Finally, there are classes whose existence may not be known. These are referred to as the \textit{unknown unknowns}. 

ZSL attempts to identify samples of \textit{known unknowns} during testing but does not study \textit{unknown unknowns}. ZSL considers  \textit{known unknowns} as unseen during training but known through the available high-level semantic descriptions. The goal of ZSL is to associate \textit{known knowns} to \textit{known unknowns} by taking advantage of the semantic descriptions available for both groups and use this association during testing to classify samples of \textit{known unknowns}. OSC acknowledges the reality of both \textit{known unknowns} and \textit{unknown unknowns} but deals with these classes only as far as they interfere in the classification of \textit{known knowns}. In other words, the critical issue in OSC is to decide whether a sample should be classified or not. Neither OSC nor ZSL addresses \textit{unknown unknowns} or tries to discover them. Therefore, NEL not only encompasses the OSC and ZSL frameworks but goes well beyond them by aiming to identify, model, and recover both \textit{known unknowns} and \textit{unknown unknowns} while simultaneously classifying \textit{known knowns} as accurately as possible.


Much of the early work in NEL employs Gaussian mixture model (GMM) and its infinite counterpart ( IGMM) \cite{akova2010machine,dundar2012bayesian,akova2012self,zhang2016bayesian}. Both GMM and IGMM are highly restrictive in that they fit each class distribution by a single Gaussian component. When classes with skewed or multi-modal distributions emerge, this limitation leads to the imprecise modeling of class distributions. Although IGMM offers additional flexibility allowing accurate estimation of the probability density function by generating an arbitrarily large number of Gaussian components, IGMM cannot address the problem of one-to-many matching between classes and components. 

In this paper, we introduce a new NEL algorithm that builds on a highly flexible  I$^{2}$GMM data model \cite{yerebakan2014infinite} and aims to unify all learning tasks by performing classification, class discovery, modeling, and recovery simultaneously. I$^{2}$GMM is a two-layer non-parametric Gaussian mixture model in which the lower layer estimates the density of the overall dataset by clustering individual data points to components and the upper layer associates components with classes to allow for recovery of class distributions. Thanks to the arbitrarily large number of components modeling each class distribution, highly flexible class distributions can be generated by I$^{2}$GMM making this method nonparametric not only regarding the number of classes but in terms of their shapes as well. Problems from two different scientific domains motivated our research. 

\textit{Planetary exploration}: Discovery of rare geological classes (phases) on Mars surface is essential as the minerals discovered (or yet to be discovered) serve as direct environmental indicators of the geochemistry of water on the planet surface. In addition to the limited spatial extension characterizing these phases, identification of rare minerals is further complicated when the phases are part of mineral assemblages, were formed as products of alteration of dominant minerals, or exist in compositionally stratified terrains. Rare phases on Mars can be considered as \textit{known unknowns} as the reference spectral signatures of these classes are known through laboratory measurements of corresponding samples found on Earth. However, the actual spectra extracted from Mars images can deviate significantly from the known signatures due to noise and artifacts. 

\textit{Flow cytometry (FC)}: FC is a single-cell screening, analysis, and sorting technology widely employed in research and clinical immunology, hematology, and oncology. The power of FC lies in its ability to quantify phenotypic characteristics of individual cells in a high-throughput manner.  Although the characteristics of cell populations (classes) present in normal samples (for instance, in blood or bone marrow), are generally known, the number of cell types and their relative proportions could be substantially different in anomalous (often diagnostically relevant) samples. The cellular phenotypes in anomalous samples can be both \textit{known unknowns}, e.g., minimal residual disease cells, and \textit{unknown unknowns}, e.g., a cancer phenotype emerging due to failed chemotherapy.

\section{Doubly Nonparametric Gaussian Mixture Model for NEL }
\label{sec:i2gmm}



I$^2$GMM models each cluster by IGMM and creates dependency across all clusters using a global DP to model the base distributions of local DPs. This two-layer architecture of I$^2$GMM allows for modeling of non-Gaussian class distributions since each class data can be represented by an arbitrarily large number of components in the lower layer. The global DP in the upper layer establishes the association between components and classes, which also allows for information sharing between and within classes.  
The generative model for I$^2$GMM is given by
\begin{equation} \label{eq:i2gmm_generative}
\begin{split}
H&=NIW(\mu,\Sigma \vert \mu_0,\Psi_0,\kappa_0,m) =N(\mu \vert \mu_0,\kappa_0^{-1}\Sigma) W^{-1} (\Sigma \vert \Psi_0,m)\\
G&\sim DP(\gamma H), \quad (\mu_k,\Sigma_k)=\theta_k \sim G, \quad H_k=N(\mu_k,\kappa_1^{-1} \Sigma_k) \\
G_k&\sim DP(\alpha H_k), \quad \mu_{kl}\sim G_k, \quad x_{kl_i}\sim N(\mu_{kl},\Sigma_k)
\end{split}
\end{equation}
 
In non-exhaustive learning, two types of discrepancies between labeled and unlabeled data can occur. The first source of disparity arises when an unlabeled data-point originates from a yet unobserved component of a known class with a multi-mode distribution or from the tail-end of a skewed distribution. The second source of error arises when an unlabeled data-point originates from an unknown class. Thus, apart from model flexibility, the other critical aspect of non-exhaustive learning is the reconciliation of these differences between labeled and unlabeled data by recovering as much information as possible from the underlying data model. I$^2$GMM has hyperparameters that control the shape of components, the scattering of class centers around the data mean, the scattering of component centers around their class means, as well as class and component sizes. These parameters define the underlying data model. The compromise between labeled and unlabeled data can be more readily reached if the model is modified to allow for learning of these hyperparameters using both labeled and unlabeled data. I$^2$GMM offers excellent flexibility for modeling datasets with an unknown number of classes where classes can have continuous arbitrary distributions. However, tuning or optimizing the values of the hyperparameters using a limited set of labeled data compromises this flexibility and yields a model that may not fit unlabeled data well.


In the Bayesian framework, it is a common practice to distribute uncertainty surrounding hyperparameters across multiple layers by treating hyperparameters as variables. An additional layer makes the model less sensitive to changes in the values of the hyperparameters. However, in a purely unsupervised setting, such a strategy significantly expands the state space, and consequently, convergence to the target distribution becomes more of a challenge. In the non-exhaustive setting presence of labeled data may help eliminate a significant portion of the potential modes the sampler can converge. However, if a large number of classes are missing and/or labeled data from existing classes are not representative of their underlying distributions some of the most promising modes might be eliminated when the labeled dataset is emphasized too strongly during model inference. We propose an adaptive I$^2$GMM that is designed as a trade-off between the model being too flexible yet uninformative vs. too restrictive and unaccommodating. 

In order to find this balanced accommodation, we first modified the generative model of I$^2$GMM by creating an additional layer in the Bayesian hierarchy that treats the most data-sensitive hyper-parameters of the model $\{\mu_{0}$,$\Psi_{0}$,$\kappa_{0}$, $\kappa_{1}\}$ as variables. Then, we introduced a weighted posterior estimation technique to infer hyper-parameters without sacrificing much from the model flexibility. During the classification process, we infer class indicator variables for unlabeled data and component indicator variables for both labeled and unlabeled data by a restricted Gibbs sampler that also takes into account potential overlaps across classes in the feature space. This adaptive version of I$^2$GMM developed for NEL is referred to as AI$^2$GMM in the following sections. Technical details are provided in the Appendix.

\section{Experiments}

We performed experiments with simulated, benchmark, and real-world datasets to validate the performance of AI$^2$GMM for non-exhaustive learning. Due to space limitations, this section focuses on experiments with simulated and real-world datasets. Results of benchmark experiments including the comparison of run times and information about experimental design are included in the Appendix \ref{apdx:experiment}.

\begin{figure}[ht]
	\begin{subfigure}[t]{0.47\textwidth}
		\includegraphics[width=\textwidth]{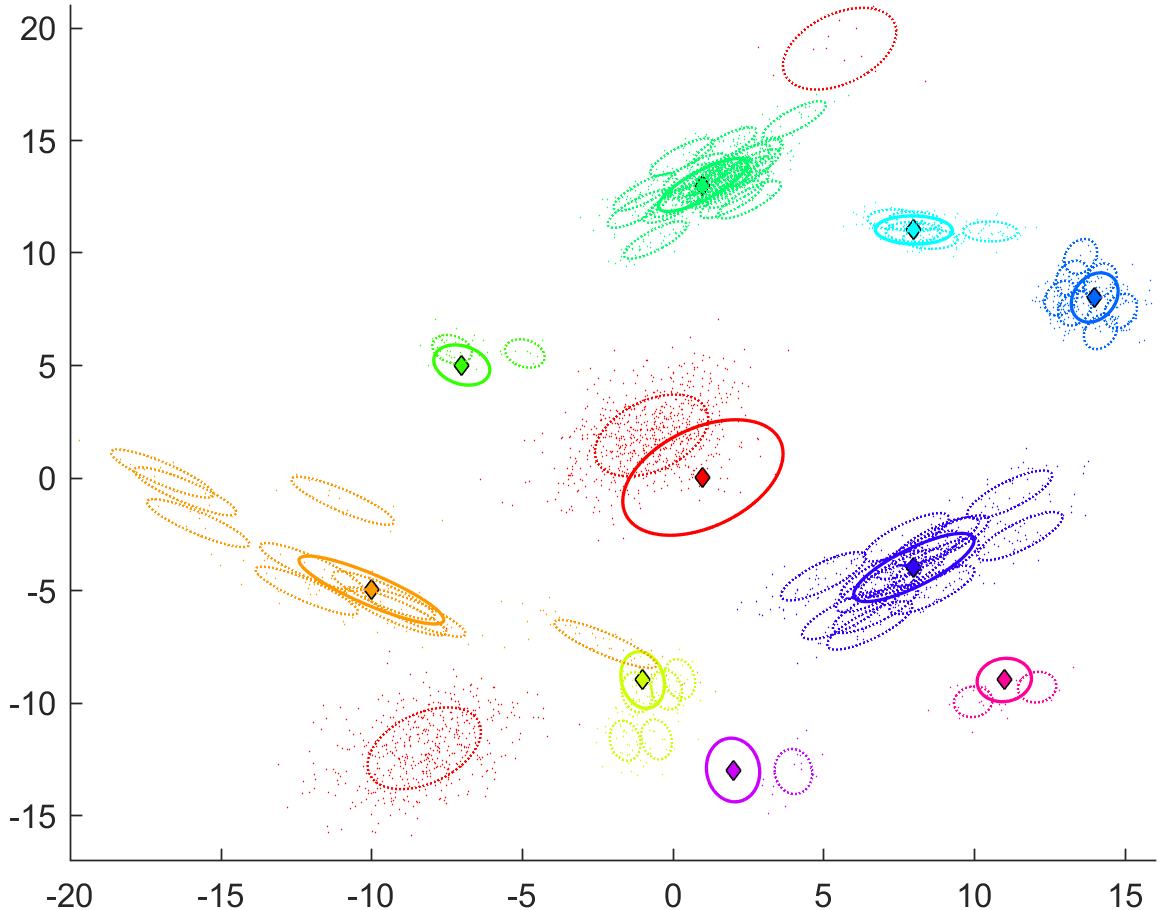}
		\caption{Scatter plot of the simulated data. Each class data is shown by a different color. Contours with solid lines indicate class-specific distributions. Contours with dashed lines indicate component distributions. } \label{fig:synthetic_draw}
    \end{subfigure}
	\begin{subfigure}[t]{0.47\textwidth}
	\includegraphics[width=\textwidth]{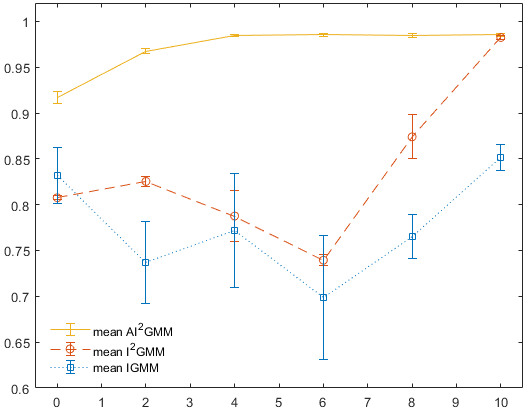}
        \caption{Mean $F_{1}$ scores as a function of the number of observed classes in the labeled data.}
        \label{fig:synthetic_f1_vague}
    \end{subfigure}
    \caption{Simulated data experiment comparing $AI^{2}GMM$, $I^{2}GMM$, and $IGMM$.}
\end{figure}

We compared the performance of AI$^2$GMM against IGMM and I$^2$GMM. IGMM and I$^2$GMM have been previously used in the literature for unsupervised learning problems. Here, we used a restricted Gibbs sampler scheme to tailor these models for NEL. In IGMM labeled points are assigned to their class of origin without sampling to preserve class composition for all observed classes. This approach is in agreement with the previous use of IGMM for NEL \cite{dundar2012bayesian,akova2012self,zhang2016bayesian}. For I$^2$GMM we used the same restricted sampling strategy we adopted for AI$^2$GMM. Briefly, we introduced two new versions of I$^2$GMM for non-exhaustive learning that both utilize the same restricted Gibbs sampler but differ in the way hyper-parameters are treated.  I$^2$GMM fixes hyper-parameters to vague values whereas  AI$^2$GMM dynamically estimates them using the combination of labeled and unlabeled data.  We employ the mean $F_{1}$ score detailed in the Appendix \ref{eq:mean_f1} to evaluate the performance of all algorithms.

In the simulated experiment we use a two layer Gaussian mixture model similar to I$^2$GMM to generate a 2D simulated data. The generated dataset is shown in Figure \ref{fig:synthetic_draw}. Classes with varying sizes exhibit multi-mode patterns with highly unbalanced component sizes, which makes this a challenging dataset for clustering.
Results for the synthetic data are shown in Figure \ref{fig:synthetic_f1_vague}.

In the first real world experiment we used a CRISM image (image id=FRT00009971) \cite{crism}. The image contains around 280K pixels where each pixel is a spectral vector of reflectance from a surface. About one-third of the image pixels are labeled by one of the six known mineral classes. We use 20$\%$ of the labeled data, and we treat the remaining 80$\%$ plus all the pixels from unlabeled regions as our unlabeled data. We first evaluate the three competing algorithms in terms of their classification performance of the labeled pixels not used in the training. IGMM achieves an $F_{1}$ score of 0.4 which significantly falls below the  scores achieved by the two other algorithms (I$^2$GMM=0.77, AI$^2$GMM=0.78). Next, we evaluate the algorithms by visually inspecting the mean vector of each newly generated class. Mean vectors represent spectral signatures of their corresponding classes. Although many of the new classes generated by all three techniques were different spectral variants of the known classes, we identified one pattern that has a distinct feature at the 2.1 micron range; a feature that does not exist in any of the known classes discovered to date. This pattern has recently become the subject of our collaborations with planetary scientists and has been confirmed to be a previously unknown spectral artifact that mimics the characteristics of real mineral absorptions \cite{ellenartefact}. The detection of this artifact has significant implications for the exploration of water and potential microbial life on Mars because it suggests a major reinterpretation of previously reported perchlorate detections in the literature \cite{ojha2015spectral}. Perchlorate is a mineral phase that is believed to offer important cues for a possible habitat for microbial life on Mars. 

In our second real-world data experiment, we used two flow cytometry samples from the EQAPOL project \cite{qiu2015computational}. The first sample contained 280K white-blood cells of a common phenotype, which we treat as the known class. The second sample included cells belonging to two very rare classes in addition to approximately the same number of cells of the abundant phenotype as in the first sample. The ratio of rare cells to the total number of cells in the second sample was around 0.001. We used all the cells from the first sample as our labeled data and all the cells from the second sample as our unlabeled with the three competing algorithms. IGMM recovered two rare classes with $F_{1}$ scores of 0.54 and 0.86 but it generated several spurious classes, which resulted in a poor $F_{1}$ score for the observed class. On the other hand, both I$^2$GMM and AI$^2$GMM not only recovered two rare classes with much higher $F_{1}$ scores (I$^2$GMM=0.78/0.90, AI$^2$GMM=0.64/0.89)  but also classified samples of the observed class with a relatively high $F_{1}$ scores (I$^2$GMM=0.89, AI$^2$GMM=0.87).

The experiments utilizing the simulated input, the benchmark data, and the real-world examples suggest that there is a strong trend that favors  I$^2$GMM and AI$^2$GMM over IGMM under all settings. The result implies the need for more flexible class distribution models. The results also suggest that estimating hyper-parameters using both labeled and unlabeled datasets offers AI$^2$GMM a significant advantage over I$^2$GMM in non-exhaustive settings. Restricting the Gibbs sampler while the model is constrained by a fixed set of vague hyper-parameters creates clustering configurations that do not conform well with the constrained model in I$^2$GMM. Under these constrained settings the model favors existing classes over new ones. This effect is more evident when the number of observed classes is small. When most or all of the classes are observed I$^2$GMM can still achieve a moderately high $F_{1}$ score without generating new classes and thus the effect of the model's limitation becomes negligible on the overall performance.

\pagebreak

\bibliographystyle{plain}
\bibliography{nips_ws18}

\pagebreak

\section*{Appendix}

\section*{Doubly Nonparametric Gaussian Mixture Model (\texorpdfstring{I$^2$GMM}{I2GMM})}
\label{sec:i2gmm_apdx}

    ~ 

In a Gaussian Mixture Model (GMM) each cluster is represented by a single Gaussian distribution characterized by its mean and covariance. In the finite GMM, the number of Gaussian components is fixed. The infinite version of GMM (IGMM) is obtained by defining a Dirichlet process (DP) prior over the components. The DP prior serves as a distribution over the Gaussian distributions. Its base distribution acts as a Bayesian prior over the mean vectors and covariance matrices while its concentration parameter models the number of components and their sizes. IGMM is an improvement over its finite version as the number of components can be inferred directly from the data. However, the core modeling aspect of GMM does not change with IGMM as each cluster data is still modeled by a single Gaussian component. This creates a problem when fitting IGMM onto dataset with multi-mode and skewed cluster distributions. In an attempt to estimate the density of the overall data more accurately, IGMM generates multiple Gaussian components for such clusters. However, more accurate density estimation does not necessarily translate into improved clustering performance with IGMM due to inherent one-to-one association imposed between components and clusters. 

I$^2$GMM is introduced to address this limitation of IGMM. I$^2$GMM models each cluster by IGMM and creates dependency across all clusters using a global DP to model the base distributions of local DPs. This two-layer architecture of I$^2$GMM allows modeling of non-Gaussian cluster shapes because each cluster data can be modeled by an arbitrarily large number of components in the lower layer. The global DP in the upper layer establishes the association between components and clusters, which also allows for information sharing across clusters and their components. 

The generative model for I$^2$GMM is given by
\begin{equation} \label{eq:i2gmm_generative_apdx}
\begin{split}
H&=NIW(\mu,\Sigma \vert \mu_0,\Psi_0,\kappa_0,m)\\
   &=N(\mu \vert \mu_0,\kappa_0^{-1}\Sigma) W^{-1} (\Sigma \vert \Psi_0,m)\\
       G&\sim DP(\gamma H)\\
(\mu_k,\Sigma_k)&=\theta_k \sim G \\       H_k&=N(\mu_k,\kappa_1^{-1} \Sigma_k) \\
G_k&\sim DP(\alpha H_k)\\
\mu_{kl}&\sim G_k\\
x_{kl_i}&\sim N(\mu_{kl},\Sigma_k)
\end{split}
\end{equation}
Based on this generative model, a global Dirichlet Process is defined with the base distribution $H$ and the concentration parameter $\gamma$.
$H$ is a bivariate Normal $\times$ Inverse Wishart distribution (NIW) with hyperparameters $\{\mu_0, \Psi_0, \kappa_0, m\}$.
To generate data points, we first draw a discrete mixing measure $G$ from $DP(\gamma H)$. 
Then the cluster centers $\mu_k$ and covariances $\Sigma_k$ are sampled from $G$.
Next, we define a local DPM with base distribution $H_k$ and concentration parameter $\alpha$ for each cluster generated by the global DPM, where $H_k$ is defined by a Gaussian centered at $\mu_k$ with covariance $\kappa_1^{-1}\Sigma_k$.
Then a cluster-specific discrete mixing measure $G_k$ is drawn from the local DPM, and a component with its mean vector drawn from $G_k$ is generated. All components generated this way share the same cluster specific covariance matrix  $\Sigma_k$. Finally data points $x_{kl_i}$ in cluster $k$ and component $l$ are generated from the component with center $\mu_{kl}$ and covariance $\Sigma_k$. The inference is performed by a collapsed Gibbs sampler as discussed in ~\cite{yerebakan2014infinite}.

\section*{Adaptive \texorpdfstring{I$^2$GMM}{I2GMM} for NEL}
\label{subsec:AI2GMM}
I$^2$GMM offers excellent flexibility for modeling dataset with an unknown number of classes where classes can have continuous arbitrary distributions. However, tuning or optimizing the values of the hyperparameters based on a limited set of labeled data compromises this flexibility and yields a model that may not fit unlabeled data well. 

In the Bayesian context, it is a common practice to distribute uncertainty surrounding hyperparameters across multiple layers by treating hyperparameters as variables. An additional layer makes the model less sensitive to changes in the values of the hyperparameters. However, in a purely unsupervised setting, such a strategy significantly expands the state space and convergence to the target distribution becomes more of a challenge. In the non-exhaustive setting presence of labeled data may help eliminate a significant portion of the potential modes the sampler can converge. However, if a large number of classes are missing and/or labeled data from existing classes are not representative of their underlying distributions, some of the most promising modes can be eliminated. This may also happen if the labeled dataset is given too much emphasis during model inference. In this section, we discuss the formulation of an adaptive I$^2$GMM that was designed to find a balance between the model being too flexible yet uninformative vs. too restrictive and unaccommodating. 

Toward achieving this end, we first modify the generative model of I$^2$GMM by creating an additional layer in the Bayesian hierarchy that treats the most data-sensitive hyper-parameters of the model $\{\mu_{0}$,$\Psi_{0}$,$\kappa_{0}$, $\kappa_{1}\}$ as variables. Then, we implement a weighted posterior estimation technique to infer hyper-parameters without sacrificing much from model flexibility. Finally, we infer class indicator variables for unlabeled data and component indicator variables for both labeled and unlabeled data by a restricted Gibbs sampler that also takes into account potential overlaps across classes in the feature space. We define a Normal prior over $\mu_0$, a Wishart prior over $\Psi_0$, Gamma priors over $\kappa_0$ and $\kappa_1$.  These prior distributions are selected for conjugacy.
\begin{equation} \label{eq:ai2gmm_generative}
\begin{aligned}
\mu_0 & \sim N \left(\mu_p,(\Psi_0c_1)^{-1} \right), \ 
\Psi_0 \sim W(\Sigma_0, c_2), \\ 
\kappa_0 & \sim \Gamma \left( \alpha_0, \beta_0 \right), \ 
\kappa_1 \sim \Gamma \left( \alpha_1, \beta_1 \right) \\
\end{aligned}
\end{equation}

\subsection*{Hyper-parameter Inference}

Note that there is no closed form solution for the posterior predictive distributions of the hyper-parameters. We resort to point estimation techniques for these hyper-parameters. To reduce the influence of spurious small classes on the globally shared hyper-parameters we incorporate weights proportional to the size of each class. 
\begin{equation}
f(\theta|*) = p(\theta) \prod_{k \in C}{l_k^{ \frac{N_k}{N}}}
\end{equation}
where $\theta$ is a generic hyper-parameter, $p(\theta)$ is the prior for the hyper-parameter $\theta$, $N$ is the number of all data points, $N_k$ is the number of data points assigned to class $k$, and $C$ is the set of classes.

The weighted posteriors conditioned on the point estimates of hidden variables are given in (\ref{eq:hyper_posterior}). We use the $\hat{}$ notation to distinguish estimates from variables. We also use $*$ to indicate conditioning on the current configuration of the data points and estimates of all other hidden variables.
\begin{equation}\label{eq:hyper_posterior}
\begin{aligned}
p(\mu_0|*) &\propto N(\mu_0|\mu_p,(\hat{\Psi}_0 c_1 )^{-1} ) \prod_{k \in C}{N(\hat{\mu}_k | \mu_0,\hat{\Sigma}_k \hat{\kappa}_0^{-1} )^{\frac{N_k}{N}}}\\
p(\Psi_0|*) &\propto W(\Psi_0|\Sigma_0,c_2 )N(\hat{\mu}_0|\mu_p,(\Psi_0 c_1 )^{-1} ) \\
&\quad \prod_{k \in C}{W^{-1} (\hat{\Sigma}_k|\Psi_0,m)^{\frac{N_k}{N}}} \\
p(\kappa_0|*) &\propto \Gamma(\kappa_0 |\alpha_0,\beta_0 ) \prod_{k\in C}{N(\hat{\mu}_k |\hat{\mu}_0,\hat{\Sigma}_k \kappa_0^{-1} )^{\frac{N_k}{N}}}\\
p(\kappa_1|*) &\propto \Gamma(\kappa_1 |\alpha_1,\beta_1 ) \prod_{k\in C}{ \prod_{l: c_l=k}{N(\hat{\mu}_{kl} |\hat{\mu}_k,\hat{\Sigma}_k \kappa_1^{-1} )}^{\frac{N_{kl}}{N}} }\\
\end{aligned}
\end{equation}

Here, $N_{kl}$ is the number of data points assigned to component $l$ in class $k$, $c_l$ is the class indicator variable for component $l$. The point estimates of the hyper-parameters are obtained by maximizing the corresponding posterior distributions.
\begin{equation} \label{eq:hyper_estimation}
\begin{aligned}
\hat{\mu}_0 &= \frac{\mu_p \hat{\Psi}_0 c_1+ \frac{\hat{\kappa}_0}{N} \sum_{k \in C}{N_k \hat{\mu}_k \hat{\Sigma}_k^{-1}}}{c_1 \hat{\Psi}_0+\frac{\hat{\kappa}_0}{N} \sum_{k \in C}{N_k \hat{\Sigma}_k^{-1}}} \\
\hat{\Psi}_0 &= \frac{c_2-d+m}{\Sigma_0^{-1}+c_1 (\hat{\mu}_0-\mu_p ) (\hat{\mu}_0-\mu_p )^T+ \frac{1}{N} \sum_{k \in C}{N_k \hat{\Sigma}_k^{-1}}} \\
\hat{\kappa}_0 &= \frac{2(\alpha_0-1)+d}{2\beta_0+ \frac{1}{N} \sum_{k}{N_k (\hat{\mu}_k-\hat{\mu}_0 )^T \hat{\Sigma}_k^{-1} (\hat{\mu}_k-\hat{\mu}_0 ) } } \\
\hat{\kappa}_1 &= \frac{2(\alpha_1-1)+d}{2\beta_1+ \frac{1}{N} \sum_{k}{ \sum_{l:c_l=k}{N_{kl} (\hat{\mu}_{kl}-\hat{\mu}_k )^T \hat{\Sigma}_k^{-1} (\hat{\mu}_{kl}-\hat{\mu}_k ) } }} \\
\end{aligned}
\end{equation}

where $d$ is the number of features of the dataset.
Note that these point estimates also depends on the hidden variables $\mu_{kl}$, $\mu_k$, $\Sigma_k$, which are estimated as follows. The weighted posterior for these variables conditioned on the point estimates of hyper-parameters and other hidden variables are given in (\ref{eq:hyper_posterior_hidden}).

\begin{equation}\label{eq:hyper_posterior_hidden}
\begin{aligned}
p(\Sigma_k |*) &\propto IW(\Sigma_k |\hat{\Psi}_0,m)N(\hat{\mu}_k |\hat{\mu}_0,\Sigma_k \hat{\kappa}_0^{-1} ) \\
& \quad \prod_{l:c_l=k}{\left( N(\hat{\mu}_{kl} |\hat{\mu}_k,\Sigma_k \hat{\kappa}_1^{-1} ) W(S_{kl} |\Sigma_k,N_{kl}-1) \right)^{\frac{N_k}{N}}} \\
p(\mu_k |*) &\propto N(\mu_k |\hat{\mu}_0,\hat{\Sigma}_k \hat{\kappa}_0^{-1} ) \prod_{l:c_l=k}{N(\hat{\mu}_{kl} |\mu_k,\hat{\Sigma}_k \hat{\kappa}_1^{-1} )^{\frac{N_{kl}}{N_k}}} \\
p(\mu_{kl} |*) &\propto N(\mu_{kl} |\hat{\mu}_k,\hat{\Sigma}_k \hat{\kappa}_1^{-1} )N(\bar{x}_{kl}|\mu_{kl}, \frac{\hat{\Sigma}_k}{N_{kl}} ) \\
\end{aligned}
\end{equation}

where $\bar{x}_{kl}$ and $S_{kl} = \sum_{i=1}^{N_{kl}}{(x_{kl_i} - \bar{x}_{kl})(x_{kl_i} - \bar{x}_{kl})^T}$ are the sample mean and scatter matrix for component  $l$ in class $k$.
The point estimates of these hidden variables are obtained by maximizing their corresponding posterior distributions.
\begin{equation} \label{eq:hyper_estimation_hidden}
\begin{aligned}
\hat{\Sigma}_k &= \frac{\hat{\Psi}_0+\hat{\kappa}_0 (\hat{\mu}_k-\hat{\mu}_0) (\hat{\mu}_k-\hat{\mu}_0 )^T + \frac{1}{N_k} \sum_{l:c_l=k}{N_{kl} S_k} }{m+d+2+ \frac{1}{N_k} \sum_{l:c_l=k}{N_{kl}^2} } \\
S_k &= \hat{\kappa}_1 (\hat{\mu}_{kl}-\hat{\mu}_k ) (\hat{\mu}_{kl}-\hat{\mu}_k )^T + S_{kl}\\
\hat{\mu}_k &= \frac{\hat{\mu}_0 \hat{\kappa}_0+\frac{\hat{\kappa}_1}{N_k}\sum_{l:c_l=k}{N_{kl}\hat{\mu}_{kl} }}{\hat{\kappa}_0+\hat{\kappa}_1} \\
\hat{\mu}_{kl} &= \frac{\hat{\mu}_k \hat{\kappa}_1+\bar{x}_{kl} N_{kl}}{\hat{\kappa}_1+N_{kl} }
\end{aligned}
\end{equation}

\subsection*{Restricted Gibbs sampler for Inference}
\label{subsec:restrict_gibbs}

Given an unlabeled dataset $X_{u}$ along with a labeled dataset $X_\ell$ and its corresponding label set $C_\ell$, a restricted Gibbs sampler is implemented to preserve the class composition of the labeled data during inference. For the points in the unlabeled dataset the class indicators $C_{u}$ and component indicators $T_{u}$ are unknown and both need to be inferred. For the data points in the labeled set, the class indicators $C_{\ell}$ are known, but the component indicators $T_{\ell}$ are unknown and need to be inferred. 

As the class labels for labeled data points are available,  a straightforward restriction would be to assign a component containing labeled data points to its corresponding class without considering other classes. Although such an approach could be useful for datasets with well-separated class distributions, it creates additional problems when classes overlap, or some of the classes exhibit heavy-tailed distributions. Assigning a component located in a region of the feature space that overlaps with other class distributions to one of the observed classes severely limits the modeling capacity of A$I^2$GMM. Many unlabeled data points from unobserved classes, which happen to be assigned to the same component as labeled data, would be incorrectly attributed to the same observed class if such a restriction were to be imposed. 

We tackle this potential class overlap problem by executing a pre-inference stage where we infer component indicators for all labeled data points and identify a fixed proportion of them from each class as potential outliers based on their class conditional likelihood values. During actual inference, each Gibbs scan includes hyper-parameter estimation followed by a sampling of component and class indicators, respectively. Component indicators for each data point are sampled without any restrictions. When sampling class-indicators for each component, we impose the following restriction. All components containing non-outlier labeled points are assigned to their respective classes without sampling. Conversely, all components composed of only unlabeled points and outlier labeled points are assigned to classes by sampling. This restricted Gibbs sampler not only preserves class composition of labeled data points but also allow for joint classification and clustering of unlabeled data points. 

Three possible scenarios exist for unlabeled data points.  \textit{1. Standard classification} occurs when unlabeled data points share the same observed component with labeled data points. \textit{2. Classification with component discovery} occurs when unlabeled data points end up in a component of an observed class with no labeled data points. \textit{3. New class discovery} occurs when unlabeled data points end up in a component such that no labeled data points are present there or across all components of its corresponding class.  

\section{Experiments}

We performed experiments with simulated, benchmark, and real-world datasets to validate the performance of AI$^2$GMM for non-exhaustive learning. For each dataset, we set aside twenty percent of all available points as labeled data. However, we do not use the labels in the labeled dataset all at once. To create a partially observed labeled dataset we start with zero observed classes, i.e., no labels are used, and gradually increase the number of observed classes by two starting with the most abundant classes, until all of the available classes for each dataset are represented in the labeled dataset. The case with no observed classes in the labeled data corresponds to a fully unsupervised setting, i.e., clustering, whereas the case with all classes observed corresponds to an entirely supervised setting, i.e., classification. All cases in between are considered to be non-exhaustive learning.  


All features in each dataset are normalized to have zero mean and unit variance. Fixed sets of vague hyper-parameters are used for both IGMM and I$^2$GMM. For IGMM: $\mu_0 = \textbf{0}$, $\Psi_0 = \textbf{I}$, $\kappa_0 = 0.1$, $m=d+2$.  For I$^2$GMM: $\mu_0=\textbf{0}$, $\Psi_0=\textbf{I}$, $\kappa_0=0.1$, $\kappa_1=0.5$, $m=d+2$ where $d$ is the number of features, $\alpha=1$, $\gamma=1$ are used for all datasets. For AI$^2$GMM we select vague values for $c_1 = 0.1$, $c_2=d+2$, $\beta_0=2d$, $\alpha_0 = 0.2d+1$,$\beta_1=2d$, $\alpha_1 = d+1$. For both I$^2$GMM and AI$^2$GMM we consider 20\% of labeled data from each observed class as outliers.  The number of Gibbs sweeps is set to 1000 in all three algorithms. For the preinference stage in AI$^2$GMM we used 100 Gibbs sweeps.
%
%
%
%
We use the mean $F_{1}$ score to evaluate the performance of all algorithms, which is computed as described below. Given the ground truth labels and predicted labels for the unlabeled data we build a confusion matrix  $C$ of $K$ rows and $L$ columns, where $K$ is the number of ground truth classes and $L$ is the number of predicted classes. We assume that rows $1$ through $M$ of $C$ are corresponding to the $M$ observed classes and columns $1$ through $M$ are corresponding to predicted classes. Mean $F_1$ score is computed as $F_{1}=\frac{1}{K} \sum_{k=1}^{K}{F_{1}^{k}}$  where $F_{1}^{k}$ is evaluated differently for observed and unobserved classes
\begin{equation} \label{eq:mean_f1}
 F_{1}^{k} = \left\{
\begin{aligned}
	\frac{2 c_{kk}}{\sum_{i=1}^{K}{c_{ik}} + \sum_{j=1}^{L}{c_{kj}}} \textrm{if k is observed} \\
    \frac{2 \max{(c_{kM+1}, ..., c_{kL})}}{\sum_{i=1}^{K}{c_{il}} + \sum_{j=1}^{L}{c_{kj}}} \textrm{if k is unobserved}
\end{aligned}
\right.
\end{equation}
where $c_{ij}$ is the element for the $ith$ row and $jth$ column of $C$ and $l$ is the index returned by $\max$. Each experiment is repeated five times and average $F_{1}$ scores are reported in all experiments to account for the stochastic nature of inference.


\subsection{Experiments with Simulated Data}
We use a two layer Gaussian mixture model similar to I$^2$GMM to generate a 2D simulated data with multi-mode class distributions. For illustration purposes  we fixed the number of classes to 10, the total number of components to 70, and the total number of points to 3000. We sample the number of components within each class and the number of data points within each component according to a Dirichlet prior with parameters set to $\gamma=1$ and $\alpha=1$, respectively. Class centers $\mu_k$ and covariance matrices $\Sigma_k$ are sampled from a NIW with $\mu_0=\textbf{0}$, $\Psi_0 = \textbf{I}$, $\kappa_0 = 0.01$, $m=4$. Component centers $\mu_{kl}$ are sampled from a Normal distribution with mean $\mu_k$ and covariance matrix  $\kappa_1^{-1}\Sigma_k$ with $\kappa_1 = 0.3$. Finally data points for each component are sampled  from a Normal distribution with mean $\mu_{kl}$ and covariance matrix $\Sigma_k$. The generated dataset is shown in Figure \ref{fig:synthetic_draw}. Classes with varying sizes exhibit multi-mode patterns with highly unbalanced component sizes, which makes this a challenging dataset for clustering.
Results for the synthetic data are shown in Figure \ref{fig:synthetic_f1_vague}.
\begin{table}[ht]
\caption{F1 scores for benchmark datasets}
\label{tbl:benchmark_results}
\begin{center}
\begin{tabular}{@{}p{1.4cm}p{1.8cm}p{1cm}p{1cm}l@{}}
\hline
dataset                      & \# observed classes  & IGMM            & I$^2$GMM          & AI$^2$GMM\\ \hline
\multirow{5}{2cm}{NDDg1 $d=12$ $N=44620$}& 0        & 0.502           & 0.696 & \textbf{0.718}          \\
                              & 2        & 0.464           & 0.654          & \textbf{0.674} \\
                              & 4        & 0.497           & 0.699          & \textbf{0.734} \\
                              & 6        & 0.535           & \textbf{0.816} & 0.803 \\
                              & 7        & 0.527           & \textbf{0.788} & 0.787 \\
\multicolumn{2}{l}{Average running time}         & 18min           & 24min          & 29min \\ \hline
\multirow{5}{1.8cm}{FLC1 $d=12$ $N=69500$} & 0        & 0.776           & 0.777 & \textbf{0.796}          \\
                              & 2        & 0.758           & 0.788          & \textbf{0.811} \\
                              & 4        & 0.797           & 0.773          & \textbf{0.821} \\
                              & 6        & 0.825           & \textbf{0.872} & 0.865 \\
                              & 9        & 0.822           & 0.931          & \textbf{0.935} \\
\multicolumn{2}{l}{Average running time}         & 31min           & 35min          & 37min \\ \hline
\multirow{3}{2cm}{stemcell $d=6$ $N=9936$}     & 0        & 0.514           & 0.651          & \textbf{0.703} \\
                              & 2        & 0.487           & 0.868 & \textbf{0.874} \\
                              & 4        & 0.566           & \textbf{0.901} & 0.815 \\
\multicolumn{2}{l}{Average running time}         & 81 sec          & 84 sec          & 85 sec \\ \hline
\end{tabular}
\end{center}
\end{table}

\subsection{Experiments with Benchmark Datasets and Computational Analyses}
\label{apdx:experiment}
We used two benchmark flow cytometry data sets and one benchmark remote sensing data set to validate the proposed algorithm for NEL. \textit{FLC1} is a 12-channel multispectral airborne image containing several land cover types most of which are known to have multi-mode class distributions. \textit{NDDg1} and \textit{stemcell} are two of the flow cytometry datasets used in the FlowCAP I competition ~\cite{aghaeepour2013critical}. Both datasets exhibit skewed class distributions.

Results of these experiments are shown in Table \ref{tbl:benchmark_results}. We also include the average running time across all experiments of the same dataset for each model in Table \ref{tbl:benchmark_results}. Even though the computational complexity for $I^2GMM$ and $AI^2GMM$ are the same ($O(CN(K+L)D^3)$, where $C$ is a large constant related to number of Gibbs sweeps, $N$ is the number of data points, $K$ is the expectation of number of clusters, $L$ is the expectation of number of components per cluster, $D$ is the dimension of the dataset), $AI^2GMM$ is slightly slower as it tend to generate more clusters than $I^2GMM$, which is a desired behavior favor to discover new classes.

\end{document}